# A New Computational Schema for Euphonic Conjunctions in Sanskrit Processing

Rama N.[1] and Meenakshi LAKSHMANAN[2]

[1] Department of Computer Science, Presidency College
Chennai 600 005, India

[2] Department of Computer Science, Meenakshi College for Women
Chennai 600 024, India
and
Research Scholar, Mother Teresa Women's University
Kodaikanal 624 101, India

**Abstract**

Automated language processing is central to the drive to enable facilitated referencing of increasingly available Sanskrit E-texts. The first step towards processing Sanskrit text involves the handling of Sanskrit compound words that are an integral part of Sanskrit texts. This firstly necessitates the processing of euphonic conjunctions or *sandhi*-s, which are points in words or between words, at which adjacent letters coalesce and transform.

The ancient Sanskrit grammarian Pāṇini's codification of the Sanskrit grammar is the accepted authority in the subject. His famed *sūtra*-s or aphorisms, numbering approximately four thousand, tersely, precisely and comprehensively codify the rules of the grammar, including all the rules pertaining to *sandhi*-s.

This work presents a fresh new approach to processing *sandhi*-s in terms of a computational schema. This new computational model is based on Pāṇini's complex codification of the rules of grammar. The model has simple beginnings and is yet powerful, comprehensive and computationally lean.

## 1. Introduction

The recognition of Sanskrit as a highly phonetic language as also one with an extensively codified grammar [1], is widespread. The very name *Saṁskṛt* (Sanskrit) means "language brought to formal perfection". That the Backus-Naur Form used in the specification of formal languages, has now come to be popularly known as the Pāṇini-Backus Form [8, 9], bears ample testimony to this fact.

Sanskrit E-texts are being increasingly made available for reference in repositories such as the Göttingen Register of Electronic Texts in Indian Languages (GRETIL) [11]. Now the essential first step towards language processing of such Sanskrit E-texts is to develop efficient algorithms and tools to handle segmentation in Sanskrit compound words that are an integral part of Sanskrit texts. This firstly necessitates the processing of *sandhi*-s or euphonic conjunctions.

### 1.1 Unicode Representation

The Unicode (UTF-8) standard is what has been adopted universally for the purpose of encoding Indian language texts into digital format. The Unicode Consortium has assigned the Unicode hexadecimal range 0900 - 097F for Sanskrit characters.

All characters including the diacritical characters used to represent Sanskrit letters in E-texts are found dispersed across the Basic Latin (0000-007F), Latin-1 Supplement (0080-00FF), Latin Extended-A (0100-017F) and Latin Extended Additional (1E00 – 1EFF) Unicode ranges.

The Latin character set has been employed in this work to represent Sanskrit letters as E-text. Moreover in this paper, any Sanskrit text except the names of people is given in italics. As such, variables such as x, y and z are not italicized as per the norm.

### 1.2 The Basis of the Work

Pāṇini, the sage and scholar dated by historians in the fourth century BC or earlier, codified the rules of the Sanskrit language based on both the extant vast literature as well as the language in prevalent use at the time. His magnum opus, the *Aṣṭādhyāyī*, which literally means 'work in eight chapters', is regarded by all scholars as the ultimate authority on Sanskrit grammar. In four parts each, these eight chapters comprise nearly four thousand *sūtra*-s or aphorisms, terse statements in Sanskrit. This grammar-codification of Pāṇini is perhaps unparalleled, for it is terse and yet comprehensive, complex yet precise. Intensive study, taking recourse to authoritative commentaries authored by adroit grammarians, is required to get a grasp of the work.

Many commentaries on the *Aṣṭādhyāyī*, such as Sage Patañjali's *Mahābhāṣya* are available and held as authentic

and comprehensive. One such authoritative commentary with a neat, topic-wise classification of Pāṇini's aphorisms, is the *Siddhānta-kaumudī* [2] written in the seventeenth century by the Sanskrit grammarian, Bhaṭṭoji Dīkṣita. The most important of these aphorisms were later extracted and compiled into the *Laghu-siddhānta-kaumudī* [10] by the scholar Varadarāja.

It is accepted among Sanskrit scholars that any exploratory work on Sanskrit grammar must necessarily have the aphorisms of Pāṇini as its basis, optionally taking recourse to any of the authoritative commentaries. This work on euphonic conjunctions is also based directly on Pāṇini's aphorisms, and not on secondary or tertiary sources of information. The *Siddhānta-kaumudī* of Bhaṭṭoji Dīkṣita, famed and accepted amongst scholars as an unabridged, comprehensive compendium of the entire *Aṣṭādhyāyī*, has been studied in the original Sanskrit, and the euphonic conjunctions dealt with in it form the basis of this work. The *Laghu-siddhānta-kaumudī* was also initially consulted for insights.

### 1.3 The *Māheśvara* aphorisms - the backbone of Pāṇini's code

The *Māheśvara* aphorisms, said to have come from the beats of a special drum called '*ḍamaru*' (hourglass drum) held in the hand of Lord Maheśvara (a form of God in the Hindu pantheon), are a set of aphorisms containing the letters of the Sanskrit alphabet in a certain sequence. These aphorisms form the basis of Pāṇini's composition of his grammar aphorisms. The *Māheśvara* aphorisms are fourteen in number and are listed below:

1. *a-i-u-ṇ*
2. *ṛ-ḷ-k*
3. *e-o-ṅ*
4. *ai-au-c*
5. *ha-ya-va-ra-ṭ*
6. *la-ṇ*
7. *ña-ma-ṅa-ṇa-na-m*
8. *jha-bha-ñ*
9. *gha-ḍha-dha-ṣ*
10. *ja-ba-ga-ḍa-da-ś*
11. *kha-pha-cha-ṭha-tha-ca-ṭa-ta-v*
12. *ka-pa-y*
13. *śa-ṣa-sa-r*
14. *ha-l*

The last letter in each of the above aphorisms is only a place-holder and is not counted as an actual letter of the aphorism. The first four aphorisms list the short forms of all the vowels, while the rest list the consonants. It must be noted that the letter '*a*' added to each of the consonants is only to facilitate pronunciation and is not part of the consonant proper.

## 2. The Problem

*Sandhi*-s in Sanskrit are points in words or between words, at which adjacent letters coalesce and transform. This is a common feature of Indian languages and is particularly elaborately dealt with and used in Sanskrit. The transformations that apply are commonly categorized into four:

1. *āgama* – addition of an extra letter or set of letters
2. *ādeśa* – substitution of one or more of the letters
3. *lopa* – dropping of a letter
4. *prakṛtibhāva* – no change

(The last is considered a transformation in the language and has therefore been listed above. However, it may be ignored for practical purposes and is hence not covered in this work.)

There are close to seventy aphorisms of Pāṇini that deal with *sandhi*-s. These aphorisms lay out the rules for the above transformations, giving the conditions under which certain letters combine with certain others to give particular results.

The challenge is to develop a computational algorithm to handle the entire range of *sandhi*-s. Such a computational algorithm would be useful to generate various word forms of a given Sanskrit word through the application of *sandhi* rules. Though this task is not difficult for a scholar of Sanskrit with a thorough knowledge of the Pāṇinian system, it is certainly a computationally non-trivial task, given the complexity and number of rules.

Existing methods of *sandhi* processing, be they methods to form compound words or even to try to split them, seem to be based on a derived understanding of the functioning of euphonic conjunctions, and usually go the finite automata-HMM-artificial intelligence way [3-7, 12]. However, the present work directly codifies Pāṇini's rules as is, recognizing that Pāṇini's codification of the grammar is based on the *Māheśvara* aphorisms that in turn lay out the letters of the alphabet in a non-trivial order. This work presents one novel method of directly representing Pāṇini's *sandhi* rules. It presents, on this basis, a mathematical formulation of a new approach to solving the non-trivial problem of handling euphonic conjunctions.

## 3. The Approach

To take advantage of the ordering of letters of the alphabet given in 2.3 above, we assign values to each letter in the Sanskrit alphabet, sticking to the order in the *Māheśvara* aphorisms rather than to the commonly adopted ordering of the letters. Thus, we have the assignment of values for the letters of the alphabet shown in Table 1.

Table 1 : Values for the letters of the Sanskrit alphabet

| Letter | Value | Letter | Value | Letter | Value |
|--------|-------|--------|-------|--------|-------|
| *a*    | 1     | *l*    | 18    | *ph*   | 35    |
| *ā*    | 2     | *ñ*    | 19    | *ch*   | 36    |
| *i*    | 3     | *m*    | 20    | *ṭh*   | 37    |
| *ī*    | 4     | *ṅ*    | 21    | *th*   | 38    |
| *u*    | 5     | *n*    | 22    | *c*    | 39    |
| *ū*    | 6     | *ṇ*    | 23    | *ṭ*    | 40    |
| *ṛ*    | 7     | *jh*   | 24    | *t*    | 41    |
| *ṝ*    | 8     | *bh*   | 25    | *k*    | 42    |
| *ḷ*    | 9     | *gh*   | 26    | *p*    | 43    |
| *e*    | 10    | *ḍh*   | 27    | *ś*    | 44    |
| *o*    | 11    | *dh*   | 28    | *ṣ*    | 45    |
| *ai* | **12** | *j* | **29** | *s* | **46** |
| *au* | **13** | *b* | **30** | *h* | **47** |
| *h* | **14** | *g* | **31** | *ṁ* | **48** |
| *y* | **15** | *ḍ* | **32** | *ḥ* | **49** |
| *v* | **16** | *d* | **33** | *'* | **50** |
| *r* | **17** | *kh* | **34** | *ru* | **51** |

Further, the letters are clubbed into various types as given below:

1. vowels: 1 – 13
2. consonants: 14 – 47
3. semi-vowels: 15 – 18
4. mutes: 19 – 47
5. nasals: 19 – 23
6. non-nasal mutes: 24 - 47
7. soft consonants: 24 – 33
8. hard consonants: 34 - 46
9. column 1: 39 – 43
10. column 2: 34 – 38
11. column 3: 29 – 33
12. column 4: 24 – 28
13. sibilants: 44 – 46
14. aspirate: 14 and 47
15. *anusvāra*: 48
16. *visarga*: 49
17. *avagraha*: 50 (replacement for the first vowel)
18. *ru*: 51 (denotes the letter *r* but is handled differently)
19. gutturals: 42, 34, 31, 26, 21
20. palatals: 39, 36, 29, 24, 19
21. cerebrals: 40, 37, 32, 27, 22
22. dentals: 41, 38, 33, 28, 23
23. labials: 43, 35, 30, 25, 20

A **rule** is the name we use for letter-level conjunctions such as the following of the *savarṇadīrgha* type: $a + \bar{a} = \bar{a}$ where the symbol '+' denotes adjacency and the term on the right of the '=' symbol is the resultant term that has to be either substituted for or added to ones on the left. (In the case of this particular *sandhi*, the term on the right is the single substitute for both terms on the left.) Substituting values of letters from Table 1, this would translate into 1+2 = 2.

Each *sandhi* may have more than one governing aphorism that specifies its functioning. Each such aphorism for every *sandhi* type in turn expands into a series of 'rules' as defined above. In this work, each and every rule for each aphorism under each of the major twenty three *sandhi* types were listed. Further, an aphorism would specify if an addition, deletion or substitution would have to be made. In accordance with this, a further cataloguing of aphorisms into five categories was done.

If we denote a *sandhi* rule as x + y = z where variables x and y denote the values of single letters joining together to yield a resultant z, then we have the following categorizations depending on both the characteristics of z and on what we actually do with it:

$C_1$: replace x and y with single-letter or multi-letter z
$C_2$: replace x with single-letter or multi-letter z
$C_3$: replace y with single-letter z
$C_4$: add single-letter z
$C_5$: drop x

Table 2 gives the summary of the numbers involved in this scenario. It must be noted that in practice, some aphorisms have to be combined or handled in two different ways to yield sets of rules, and hence what may seem to be a discrepancy in the number of aphorisms shown in the table and the number of rows shown for the rules of that aphorism, is no real discrepancy at all.

As can be seen, there are close to 2500 individual rules involved, even with considering only the major *sandhi*-s. Tabulation of these rules in terms of x, y and z for the categories and then tabulation of the corresponding values as per Table 1 were done.

Table 2: Summary of the number of Sanskrit *sandhi* aphorisms and rules

| # | *Sandhi* Type | No. of *sūtra*-s | Categories | | | | | No. of Rules |
|---|---|---|---|---|---|---|---|---|
| | | | $C_1$ | $C_2$ | $C_3$ | $C_4$ | $C_5$ | |
| 1 | *yaṇādeśa* | 1 | | 74 | | | | 74 |
| 2 | *ayāyāvāvādeśa* | 4 | | 50 | | | | 50 |
| | | | | 2 | | | | 2 |
| | | | | 3 | | | | 3 |
| 3 | *guṇa* | 2 | 8 | | | | | 8 |
| | | | 18 | | | | | 18 |
| 4 | *vṛddhi* | 3 | 8 | | | | | 8 |
| | | | 18 | | | | | 18 |
| | | | 10 | | | | | 10 |
| 5 | *pararūpa* | 1 | 10 | | | | | 10 |
| 6 | *savarṇadīrgha* | 1 | 15 | | | | | 15 |
| 7 | *pūrvarūpa* | 1 | 2 | | | | | 2 |
| 8 | *avaṅādeśa* | 1 | | 13 | | | | 13 |
| 9 | *tugāgama* | 4 | | | | 13 | | 13 |
| | | | | | | 1 | | 1 |
| 10 | *jaśtva* | 2 | | 23 | | | | 23 |
| | | | | 240 | | | | 240 |
| 11 | *satva* | 2 | | 5 | | | | 5 |
| | | | | 230 | | | | 230 |
| | | | | 138 | | | | 138 |
| 12 | *anusvāra* | 5 | | 34 | | | | 34 |
| | | | | 24 | | | | 24 |
| | | | | 1 | | | | 1 |
| | | | | 3 | | | | 3 |
| 13 | *dhuḍāgama* | 2 | | | | 2 | | 2 |
| 14 | *ṅamudāgama* | 1 | | | | 195 | | 195 |
| 15 | *ścutva* | 2 | | 36 | | | | 36 |
| | | | | | 31 | | | 31 |
| 16 | *ṣṭutva* | 3 | | 31 | | | | 31 |
| | | | | | 6 | | | 6 |
| 17 | *anunāsikā* | 1 | | 160 | | | | 160 |
| 18 | *cartva* | 1 | | 312 | | | | 312 |
| 19 | *parasavarṇa* | 3 | | 29 | | | | 29 |
| | | | | 5 | | | | 5 |
| 20 | *pūrvasavarṇa* | 1 | | | 20 | | | 20 |
| 21 | *chatva* | 1 | | | 340 | | | 340 |
| 22 | *visarga* | 2 | | 13 | | | | 13 |
| | | | | 13 | | | | 13 |
| 23 | *svādi* | 5 | | | | | 66 | 66 |
| | | | | | | | 13 | 13 |
| | | | | | | | 132 | 132 |
| | | | | | | | 33 | 33 |
| | | | | | | | 33 | 33 |
| **TOTAL** | | 49 | 89 | 1439 | 397 | 211 | 277 | **2413** |

Careful observations based on a thorough understanding of the domain and classification of the input conditions, yielded the equations presented in this work.

We define the general binary operators $\oplus_1, \oplus_2, \oplus_3, \oplus_4$ and $\oplus_5$ for the categories $C_1, C_2, C_3, C_4$ and $C_5$ respectively, as follows:

$C_1$: $\oplus_1(x, y) = z = z_1$
$C_2$: $\oplus_2(x, y) = z = z_2 y$
$C_3$: $\oplus_3(x, y) = z = x z_3$
$C_4$: $\oplus_4(x, y) = z = x z_4 y$
$C_5$: $\oplus_5(x, y) = z = y$

where each of $z_1, z_2, z_3, z_4$ is to be calculated. Now we introduce the second and third subscripts for the above general operators as follows: the general operator $\oplus_{i,j}(x, y)$ is derived from $\oplus_i$ and signifies the operator applying to aphorism number j of Category $C_i$; the specialized operator $\oplus_{i,j,k}(x, y)$ is derived from the operator $\oplus_{i,j}$ and appertains to the $k^{th}$ equation for the $j^{th}$ aphorism of Category $C_i$. These two extra subscripts are necessitated by the facts that a category encompasses many aphorisms and one aphorism may itself be governed by more than one equation.

## 4. Results and Discussion

The main *sandhi* aphorisms, their brief description (Rule), the corresponding general operator and the final, specialized equations along with the domain of operation are given below in a category-wise listing. Special notations followed are:

- The equations and conditions given as operators with three subscripts are the ones that are implementable. The 'general operator' specified for each aphorism typifies the aphorism's meaning and all the conditions it becomes operative under, and provides a generalization from which the final equations are specialized. A specialized operator would thus override the 'general operator' with its own specialized conditions.
- The variable X denotes the sequence of letters culminating in x; the variable Y denotes the sequence of letters starting with y. These are used to depict special conditions that pertain to the entire word involved in the *sandhi*.
- Variables u and w represent the value of the letter occurring just before x and just after y respectively.
- [ ] are used to club domain conditions simply in order to depict the 'or' condition more clearly.

### 4.1 $C_1$ *Sandhi*-s

*guṇa sandhi*

1. *ādguṇaḥ* || 6.1.87 ||
   Rule: *a* or *ā* followed by *i, u* (short and long) -> *guṇa* letter (*e, o*) corresponding to second letter replaces both.

   General operator: $\oplus_{1,1}(x, y) = z = z_1$ when $x \in \{1, 2\}$, $y \in \{3, 4, 5, 6\}$

   $\oplus_{1,1,1}(x, y) : z_1 = 10$ when $y \in \{3, 4\}$

   $\oplus_{1,1,2}(x, y) : z_1 = 11$ when $y \in \{5, 6\}$

2. *uraṇ raparaḥ* || 1.1.51 ||
   Rule: *a* or *ā* followed by *ṛ* (short and long), *ḷ* -> *guṇa* letter (*ar, al*) corresponding to the second letter replaces both.

   General operator: $\oplus_{1,2}(x, y) = z = z_1 = z_{11} z_{12}$ when $x \in \{1, 2\}, y \in \{7, 8, 9\}$
   $\oplus_{1,2,1}(x, y) : z_{11} = 1, z_{12} = 17$ when $y \in \{7, 8\}$
   $\oplus_{1,2,2}(x, y) : z_{11} = 1, z_{12} = 18$ when $y \in \{9\}$

*vṛddhi sandhi*

3. *vṛddhireci* || 6.1.88 ||
   Rule: *a* or *ā* followed by *e, o, ai, au* -> *vṛddhi* letter (*ai, au*) corresponding to second letter replaces both.
   General operator: $\oplus_{1,3}(x, y) = z = z_1$ when $x \in \{1, 2\}$, $y \in \{10, 11, 12, 13\}$
   $\oplus_{1,3,1}(x, y) : z_1 = y + 2$ when $y \in \{10, 11\}$
   $\oplus_{1,3,1}(x, y) : z_1 = y$ when $y \in \{12, 13\}$

4. *etyedhatyūṭhsu* || 6.1.89 ||
   Rule: In all the following rules, *vṛddhi* letter (*ai, au, ār, āl*) corresponding to the beginning of second word, replaces both.
   a. *a* or *ā* followed by the prepositions *eti, edhati* -> *ai* replaces both
   b. preposition *pra* followed by *eṣaḥ, eṣya* -> *ai* replaces both
   c. word *sva* followed by *īr* -> *ai* replaces both
   d. *a* or *ā* followed by the preposition *ūh* -> *au* replaces both
   e. word *akṣa* followed by word *ūhini* -> *au* replaces both
   f. preposition *pra* followed by *ūh, ūḍh* -> *au* replaces both

   General operator: $\oplus_{1,4}(x, y) = z = z_1$ when $x \in \{1, 2\}$
   $\oplus_{1,4,1}(x, y) : z_1 = 12$ when $[y = 10, Y \in \{10+41+3, 10+28+1+41+3\}]$ or $[x = 1, y = 10, X \in \{43+17+1\}]$ or $[x = 1, y = 4, X \in \{46+16+1\}, Y \in \{4+17\}]$
   $\oplus_{1,4,2}(x, y) : z_1 = 13$ when $[y = 6, Y \in \{6+14\}]$ or $[x = 1, y = 6, X \in \{1+42+45+1\}, Y \in \{6+14+3+23+3\}]$ or $[x = 1, y = 6, X \in \{43+17+1\}, Y \in \{4+17, 4+27\}]$

5. *etyedhatyūṭhsu* || 6.1.89 ||
   Rule: In all the following rules, *vṛddhi* letter (*ai, au, ār, āl*) corresponding to the beginning of second word, replaces both.
   a. *a* followed by word *ṛta* -> *ār* replaces both
   b. preposition/words *pra, vatsara, kambala, vasana, daśa, ṛṇa* followed by the word *ṛṇa* -> *ār* replaces both

   General operator: $\oplus_{1,5}(x, y) = z = z_1 = z_{11} z_{12}$ when $x = 1$
   $\oplus_{1,5,1}(x, y) : z_{11} = 2, z_{12} = y + 10$ when $[y = 7, Y \in \{7+41+1\}]$ or $[X \in \{43+17+1, 16+1+41+46+ 1+17+1, 42+1+20+30+1+18+1, 16+1+46+1+23+ 1, 33+1+44+1, 7+22+1\}, Y \in \{7+22+1\}]$

6. *upasargādṛti dhātau* || 6.1.91 ||
   Rule: *a* or *ā* at the end of prepositions followed by *ṛ* -> *vṛddhi* letter *ār* replaces both. (The prepositions that qualify are: *pra, parā, apa, ava, upa*)

General operator: $\oplus_{1,6}(x, y) = z = z_1 = z_{11}z_{12}$ when $x \in \{1, 2\}$, $y = 7$

$\oplus_{1,6,1}(x, y) : z_{11} = 2$, $z_{12} = y + 10$ when $X \in \{43+17+1, 43+1+17+2, 1+43+1, 1+16+1, 5+43+1\}$

*pararūpa sandhi*

7. *eṅi pararūpaṁ* || 6.1.94 ||
   Rule: *a* or *ā* at the end of a preposition followed by *e* or *o* (of a verbal root) -> second letter (*e* or *o*) replaces both.
   General operator: $\oplus_{1,7}(x, y) = z = z_1 = y$ when $x \in \{1, 2\}$, $y \in \{10, 11\}$
   $\oplus_{1,7,1}(x, y) : z_1 = y$ when $x \in \{1, 2\}$, $y \in \{10, 11\}$, $X \in \{43+17+1, 43+1+17+2, 1+43+1, 1+16+1, 5+43+1\}$

*savarṇadīrgha sandhi*

8. *akaḥ savarṇe dīrghaḥ* || 6.1.101 ||
   Rule: *a, i, u, ṛ, ḷ* (short or long) followed by similar *a, i, u, ṛ, ḷ* (short or long) -> corresponding long letter replaces both.
   General operator: $\oplus_{1,8}(x, y) = z = z_1 = y$ when $1 \le x \le 9$, $1 \le y \le 9$
   All operators $\oplus_{1,8,i}$ are commutative.
   $\oplus_{1,8,1}(x, y) : z_1 = y$ when $[x \in \{1, 3, 5\}, y = x+1]$ or $[x \in \{2, 4, 6\}, y = x]$
   $\oplus_{1,8,2}(x, y) : z_1 = y + 1$ when $x \in \{1, 3, 5\}$, $y = x$
   $\oplus_{1,8,3}(x, y) : z_1 = 8$ when $x, y \in \{7, 8, 9\}$

*pūrvarūpa sandhi*

9. *eṅaḥ padāntādati* || 6.1.109 ||
   Rule: *e* or *o* followed by *a* -> first letter replaces both.
   General operator: $\oplus_{1,9}(x, y) = z = z_1 = x$ when $x \in \{10, 11\}$, $y = 1$
   $\oplus_{1,9,1}(x, y) : z_1 = x$ when $x \in \{10, 11\}$, $y = 1$

4.2 $C_2$ *Sandhi*-s

*yaṇādeśa sandhi*

1. *iko yaṇaci* || 6.1.77 ||
   Rule: *i, u, ṛ, ḷ* (short and long) followed by dissimilar vowel -> *y, v, r, l* respectively replace first letter.
   General operator: $\oplus_{2,1}(x, y) = z = z_2y$ when $3 \le x \le 9$, $y \le 13$

   $\oplus_{2,1,1}(x, y) : z_2 = 15$ when $x \in \{3, 4\}$, $y \notin \{3, 4\}$

   $\oplus_{2,1,2}(x, y) : z_2 = 16$ when $x \in \{5, 6\}$, $y \notin \{5, 6\}$

   $\oplus_{2,1,3}(x, y) : z_2 = 17$ when $x \in \{7, 8\}$, $y \notin \{7, 8, 9\}$

   $\oplus_{2,1,4}(x, y) : z_2 = 18$ when $x \in \{9\}$, $y \notin \{7, 8, 9\}$

*ayāya-avāva-ādeśa sandhi*

2. *ecoyavāyāvaḥ* || 6.1.78 ||
   Rule: *e, o* followed by *āc* -> *ay, av* replace the first respectively;
   *ai, au* followed by *ac* -> *āy, āv* replace the first respectively.
   General operator: $\oplus_{2,2}(x, y) = z = z_2y = z_{21}z_{22}y$ when $10 \le x \le 13$, $y \le 13$

   $\oplus_{2,2,1}(x, y) : z_{21} = 1$, $z_{22} = x + 5$ when $x \in \{10, 11\}$, $y \ne 1$

   $\oplus_{2,2,2}(x, y) : z_{21} = 2$, $z_{22} = x + 3$ when $x \in \{12, 13\}$

3. *vānto yi pratyaye* || 6.1.79 ||
   Rule: *o, au* followed by *y* -> *av, āv* replace the first respectively.
   General operator: $\oplus_{2,3}(x, y) = z = z_2y = z_{21}z_{22}y$ when $x \in \{11, 13\}$, $y = 15$
   $\oplus_{2,3,1}(x, y) : z_{21} = 1$, $z_{22} = x + 5$ when $x = 11$
   $\oplus_{2,3,1}(x, y) : z_{21} = 2$, $z_{22} = x + 3$ when $x = 13$

4. *kṣayyajayyau śakyārthe* || 6.1.81 ||
   *krayyastadarthe* || 6.1.82 ||
   Rule: *e* which is the end of words *kṣe, je, kre* followed by *y* -> *ay* replaces the first.
   General operator: $\oplus_{2,4}(x, y) = z = z_2y = z_{21}z_{22}y$ when $x = 10$, $y = 15$, $X \in \{42+45+10, 29+10\}$
   $\oplus_{2,4,1}(x, y) : z_{21} = 1$, $z_{22} = x + 5$

*avaṅādeśa sandhi*

5. *avaṅ sphoṭāyanasya* || 6.1.123 ||
   Rule: *o* which is the end of word *go* followed by a vowel -> '*ava*' replaces the first.
   General operator: $\oplus_{2,5}(x, y) = z = z_2y = z_{21}z_{22}z_{23}y$ when $x = 11$, $y \le 13$, $X = 31+11$
   $\oplus_{2,5,1}(x, y) : z_{21} = 1$, $z_{22} = 16$, $z_{23} = 1$

*jaśtva sandhi*

6. *jhalāṁ jaśo'nte* || 8.2.39 ||
   Rule: non-nasal mutes, sibilants, aspirate at the end of a word -> first letter replaced by corresponding column 3 letter.
   General operator: $\oplus_{2,6}(x, y) = z = z_2y$ when $24 \le x \le 47$, $y = 0$
   $\oplus_{2,6,1}(x, y) : z_2 = x + 5$ when $x \in \{24, 25, 26, 27, 28\}$
   $\oplus_{2,6,2}(x, y) : z_2 = x$ when $x \in \{29, 30, 31, 32, 33, 44, 45, 47\}$
   $\oplus_{2,6,3}(x, y) : z_2 = x - 3$ when $x = 34$
   $\oplus_{2,6,4}(x, y) : z_2 = x - 5$ when $x \in \{35, 37, 38\}$
   $\oplus_{2,6,5}(x, y) : z_2 = x - 7$ when $x = 36$
   $\oplus_{2,6,6}(x, y) : z_2 = x - 8$ when $x \in \{40, 41\}$
   $\oplus_{2,6,7}(x, y) : z_2 = x - 11$ when $x = 42$
   $\oplus_{2,6,8}(x, y) : z_2 = x - 13$ when $x = 43$
   $\oplus_{2,6,9}(x, y) : z_2 = x - 10$ when $x = 39$

*satva sandhi*

7. *samaḥ suṭi* || 8.3.5 ||
   Rule 1: word *sam* followed by affixes *kṛ, kṝ, kar, kār, kur* -> *m* of *sam* replaced with the combination *ṁs*.
   General operator: $\oplus_{2,7}(x, y) = z = z_2y = z_{21}z_{22}y$ when $x = 20$, $y = 42$, $X \in \{46+1+20\}$, $Y \in \{42+7, 42+8, 42+1+17, 42+2+17, 42+5+17\}$
   $\oplus_{2,7,1}(x, y) : z_{21} = 48$, $z_{22} = 46$

8. *pumaḥ khayyampare* || 8.3.6 ||
   Rule: word *pum* followed by column 1, column 2 which is in turn followed by a vowel, aspirate, semi-vowel or nasal -> ending *m* replaced with the combination *ṁs*.

General operator: $\oplus_{2,8}$ (x, y) = z = $z_2$y = $z_{21}z_{22}$y when x = 20, 34<=y<=43, 1<=w<=23

$\oplus_{2,8,1}$ (x, y) : $z_{21}$ = 48, $z_{22}$ = 46

9. *naśchavyapraśān* || 8.3.7 ||
   Rule: final *n* of a word except for the word *praśān*, followed by *ch, ṭh, th, c, ṭ, t* which is in turn followed by a vowel, aspirate, semi-vowel or nasal -> ending *n* replaced with the combination *ṁs*.
   General operator: $\oplus_{2,9}$ (x, y) = z = $z_2$y = $z_{21}z_{22}$y when x = 23, 36<=y<=41, 1<=w<=23, X ∉ {43+17+1+44+2+23}
   $\oplus_{2,9,1}$ (x, y) : $z_{21}$ = 48, $z_{22}$ = 46

*visarga sandhi*

10. *kharavasānayorvisarjanīyaḥ* || 8.3.15 ||
    Rule: *r* followed by hard consonant -> *r* replaced with *visarga*.
    General operator: $\oplus_{2,10}$ (x, y) = z = $z_2$y when x = 17, 34 <= y <= 46
    $\oplus_{2,10,1}$ (x, y) : $z_2$ = 49

*anusvāra sandhi*

11. *mo'nusvāraḥ* || 8.3.23 ||
    *mo rāji samaḥ kvau* || 8.3.25 ||
    Rule: *m* followed by any consonant -> *m* letter replaced by *ṁ* (*anusvāra*) (except in the case of the word *sam* being followed by the word *rāṭ*)
    General operator: $\oplus_{2,11}$ (x, y) = z = $z_2$y when x = 20, 14 <= y <= 47, X ∉ {46+1+20}, Y ∉ {17+2+40}
    $\oplus_{2,11,1}$ (x, y) : $z_2$ = 48

12. *naścāpadāntasya jhali* || 8.3.24 ||
    Rule: *n* followed by a non-nasal mute, sibilant or aspirate (not at the end of a *pada*) -> *n* replaced by *ṁ* (*anusvāra*).
    General operator: $\oplus_{2,12}$ (x, y) = z = $z_2$y when x = 23, 24 <= y <= 47
    $\oplus_{2,12,1}$ (x, y) : $z_2$ = 48

13. *he mapare vā* || 8.3.26 ||
    Rule: *m* followed by *h* which is in turn followed by *y, l,* or *v* -> the first *m* replaced by nasal *y, l, v* (i.e. *ṁy, ṁl, ṁv*) respectively.
    General operator: $\oplus_{2,13}$ (x, y) = z = $z_2$y = $z_{21}z_{22}$y when x = 20, y = 14, w ∈ {15, 16, 18}
    $\oplus_{2,13,1}$ (x, y) : $z_{21}$ = 48, $z_{22}$ = w

14. *napare naḥ* || 8.3.27 ||
    Rule: *m* followed by *h* at the end of a *pada* which is in turn followed by *n* -> *m* replaced by *n*.
    General operator: $\oplus_{2,14}$ (x, y) = z = $z_2$y when x = 20, y = 14, w = 23
    $\oplus_{2,14,1}$ (x, y) : $z_2$ = w

*visarga sandhi*

15. *visarjanīyasya saḥ* || 8.3.34 ||
    Rule: *visarga* followed by hard consonant –> *visarga* replaced with *s*.

General operator: $\oplus_{2,15}$ (x, y) = z = $z_2$y when x = 49, 34 <= y <= 46, w ∉ {44, 45, 46}

$\oplus_{2,15,1}$ (x, y) : $z_2$ = 46

*ścutva sandhi*

16. *stoḥ ścunāḥ ścuḥ* || 8.4.40 ||
    Rule: dentals, *s* followed by palatals, *ś* -> first replaced by its corresponding palatal, *ś* respectively.
    General operator: $\oplus_{2,16}$ (x, y) = z = $z_2$y when x ∈ {41, 38, 33, 28, 23, 46}, y ∈ {39, 36, 29, 24, 19, 44}
    $\oplus_{2,16,1}$ (x, y) : $z_2$ = x − 2 when x ∈ {41, 38, 46}
    $\oplus_{2,16,2}$ (x, y) : $z_2$ = x − 4 when x ∈ {33, 28, 23}

*ṣṭutva sandhi*

17. *ṣṭunāḥ ṣṭuḥ* || 8.4.41 ||
    *toḥ ṣi* || 8.4.43 ||
    Rule: [dentals, *s* followed by cerebrals] or [*s* followed by *ṣ*] -> dentals or *s* replaced by cerebrals or *ṣ* respectively.
    General operator: $\oplus_{2,17}$ (x, y) = z = $z_2$y when x ∈ {41, 38, 33, 28, 23, 46}, y ∈ {40, 37, 32, 27, 22, 45}
    $\oplus_{2,17,1}$ (x, y) : $z_2$ = x − 1 when [x = 46] or [y != 45]

*anunāsikā sandhi*

18. *yaro'nunāsike'nunāsiko vā* || 8.4.45 ||
    Rule: semi-vowels *y, v* and *l* followed by nasal -> first replaced by its corresponding nasal, *ṁy, ṁv, ṁl* respectively.
    General operator: $\oplus_{2,18}$ (x, y) = z = $z_2$y = $z_{21}z_{22}$y when x ∈ {15, 16, 18}, 19 <= y <= 23
    $\oplus_{2,18,1}$ (x, y) : $z_{21}$ = 48, $z_{22}$ = x

19. *yaro'nunāsike'nunāsiko vā* || 8.4.45 ||
    Rule: semi-vowel *r*, mutes, sibilants followed by nasal -> first replaced by its corresponding nasal.
    General operator: $\oplus_{2,19}$ (x, y) = z = $z_2$y when 17 <= x <= 46, x != 18, 19 <= y <= 23
    $\oplus_{2,19,1}$ (x, y) : $z_2$ = x when x ∈ {17, 19, 20, 21, 22, 23, 44, 45, 46}
    $\oplus_{2,19,2}$ (x, y) : $z_2$ = x − 5 when x ∈ {24, 25, 26, 27, 28}
    $\oplus_{2,19,3}$ (x, y) : $z_2$ = x − 10 when x ∈ {29, 30, 31, 32, 33}
    $\oplus_{2,19,4}$ (x, y) : $z_2$ = x − 13 when x = 34
    $\oplus_{2,19,5}$ (x, y) : $z_2$ = x − 15 when x ∈ {35, 37, 38}
    $\oplus_{2,19,6}$ (x, y) : $z_2$ = x − 17 when x = 36
    $\oplus_{2,19,7}$ (x, y) : $z_2$ = x − 18 when x ∈ {40, 41}
    $\oplus_{2,19,8}$ (x, y) : $z_2$ = x − 20 when x = 39
    $\oplus_{2,19,9}$ (x, y) : $z_2$ = x − 21 when x = 42
    $\oplus_{2,19,10}$ (x, y) : $z_2$ = x − 23 when x = 43

*jaśtva sandhi*

20. *jhalāṁ jaś jhaśi* || 8.4.53 ||
    Rule: non-nasal mutes, sibilants, aspirate followed by soft consonants (column 3, column 4) -> first replaced by corresponding column 3 letter.
    General operator: $\oplus_{2,20}$ (x, y) = z = $z_2$y when 24 <= x <= 47, 24 <= y <= 33
    $\oplus_{2,20,1}$ (x, y) : $z_2$ = x + 5 when 24 <= x <= 28

$\oplus_{2,20,2}$ (x, y) : $z_2$ = x  when x ∈ {29, 30, 31, 32, 33, 44, 45, 46, 47}

$\oplus_{2,20,3}$ (x, y) : $z_2$ = x – 3 when x = 34

$\oplus_{2,20,4}$ (x, y) : $z_2$ = x – 5 when x ∈ {35, 37, 38}

$\oplus_{2,20,5}$ (x, y) : $z_2$ = x – 7 when x = 36

$\oplus_{2,20,6}$ (x, y) : $z_2$ = x – 8 when x ∈ {40, 41}

$\oplus_{2,20,7}$ (x, y) : $z_2$ = x – 10 when x = 39

$\oplus_{2,20,8}$ (x, y) : $z_2$ = x – 11 when x = 42

$\oplus_{2,20,9}$ (x, y) : $z_2$ = x – 13 when x = 43

*cartva sandhi*

21. *khari ca* || 8.4.55 ||
    Rule: non-nasal mutes, sibilants, aspirate followed by hard consonants (column 3, column 4, sibilants) -> first replaced by its corresponding column 1 or sibilants.
    General operator: $\oplus_{2,21}$ (x, y) = z = $z_2$y   when 24 <= x <= 47, 34 <= y <= 46
    
    $\oplus_{2,21,1}$ (x, y) : $z_2$ = x + 18 when x = 25
    $\oplus_{2,21,2}$ (x, y) : $z_2$ = x + 16 when x = 26
    $\oplus_{2,21,3}$ (x, y) : $z_2$ = x + 15 when x = 24
    $\oplus_{2,21,4}$ (x, y) : $z_2$ = x + 13 when x ∈ {27, 28, 30}
    $\oplus_{2,21,5}$ (x, y) : $z_2$ = x + 11 when x = 31
    $\oplus_{2,21,6}$ (x, y) : $z_2$ = x + 10 when x = 29
    $\oplus_{2,21,7}$ (x, y) : $z_2$ = x + 8 when x ∈ {32, 33, 34, 35}
    $\oplus_{2,21,8}$ (x, y) : $z_2$ = x + 3 when x ∈ {36, 37, 38}
    $\oplus_{2,21,9}$ (x, y) : $z_2$ = x       when 39 <= x <= 47

*parasavarṇa sandhi*

22. *anusvārasya yayi parasavarṇaḥ* || 8.4.58 ||
    Rule: *anusvāra* followed by semi-vowels, mutes -> *anusvāra* replaced by the nasal equivalent of the second.
    General operator: $\oplus_{2,22}$ (x, y) = z = $z_2$y  when x = 48, 15 <= y <= 43
    
    $\oplus_{2,22,1}$ (x, y) : $z_2$ = 20  when x ∈ {16, 17}
    $\oplus_{2,22,2}$ (x, y) : $z_2$ = y   when x ∈ {15, 18, 19, 20, 21, 22, 23}
    $\oplus_{2,22,3}$ (x, y) : $z_2$ = y – 5 when 24 <= x <= 28
    $\oplus_{2,22,4}$ (x, y) : $z_2$ = y – 10 when 29 <= x <= 33
    $\oplus_{2,22,5}$ (x, y) : $z_2$ = y – 13 when x = 34
    $\oplus_{2,22,6}$ (x, y) : $z_2$ = y – 15 when x ∈ {35, 37, 38}
    $\oplus_{2,22,7}$ (x, y) : $z_2$ = y – 17 when x = 36
    $\oplus_{2,22,8}$ (x, y) : $z_2$ = y – 18 when x ∈ {40, 41}
    $\oplus_{2,22,9}$ (x, y) : $z_2$ = y – 20 when x = 39
    $\oplus_{2,22,10}$ (x, y) : $z_2$ = y – 21 when x = 42
    $\oplus_{2,22,11}$ (x, y) : $z_2$ = y – 23 when x = 43

23. *torli* || 8.4.60 ||
    Rule 1: dentals except *n* followed by *l* -> dentals replaced by *l*.
    General operator: $\oplus_{2,23}$ (x, y) = z = $z_2$y when x ∈ {41, 38, 33, 28}, y = 18
    $\oplus_{2,23,1}$ (x, y) : $z_2$ = y

24. *torli* || 8.4.60 ||
    Rule 2: *n* followed by *l* -> *n* replaced by nasal *l* (i.e. *m̐l*).
    General operator: $\oplus_{2,24}$ (x, y) = z = $z_2$y = $z_{21}z_{22}$y  when x = 23, y = 18
    $\oplus_{2,24,1}$ (x, y) : $z_{21}$ = 48, $z_{22}$ = y

### 4.3 $C_3$ *Sandhi*-s

*ścutva sandhi*

1. *stoḥ ścunāḥ ścuḥ* || 8.4.40 ||
   *śāt* || 8.4.44 ||
   Rule: [palatals followed by dentals, *s*] or [*ś* followed by *s*] -> second replaced by palatals or *ś* respectively.
   General operator: $\oplus_{3,1}$ (x, y) = z = $xz_3$  when x ∈ {39, 36, 29, 24, 19, 44}, y ∈ {41, 38, 33, 28, 23, 46}
   $\oplus_{3,1,1}$ (x, y) : $z_3$ = y – 2 when [y = 46] or [x != 44, y ∈ {41, 38}]
   $\oplus_{3,1,2}$ (x, y) : $z_3$ = y – 4 when x != 44, y ∈ {33, 28, 23}

*ṣṭutva sandhi*

2. *ṣṭunāḥ ṣṭuḥ* || 8.4.41 ||
   *na padāntāṭṭoranām* || 8.4.42 ||
   Rule: *ṣ* followed by dentals, *s* -> dentals, *s* replaced by cerebrals, *ṣ* respectively.
   General operator: $\oplus_{3,2}$ (x, y) = z = $xz_3$  when x = 45, y ∈ {41, 38, 33, 28, 23, 46}
   $\oplus_{3,2,1}$ (x, y) : $z_3$ = y – 1

*pūrvasavarṇa sandhi*

3. *jhayo ho'nyatarasyām* || 8.4.62 ||
   Rule: non-nasal mutes followed by *h* -> *h* replaced by the aspirate letter (column 4) corresponding to the first non-nasal mute.
   General operator: $\oplus_{3,3}$ (x, y) = z = $xz_3$ when 24 <= x <= 43, y = 47
   $\oplus_{3,3,1}$ (x, y) : $z_3$ = x   when 24 <= x <= 28
   $\oplus_{3,3,2}$ (x, y) : $z_3$ = x - 5 when 29 <= x <= 33
   $\oplus_{3,3,3}$ (x, y) : $z_3$ = x - 8  when x = 34
   $\oplus_{3,3,4}$ (x, y) : $z_3$ = x – 10  when x ∈ {35, 37, 38}
   $\oplus_{3,3,5}$ (x, y) : $z_3$ = x – 12  when x = 36
   $\oplus_{3,3,6}$ (x, y) : $z_3$ = x – 13  when x ∈ {40, 41}
   $\oplus_{3,3,7}$ (x, y) : $z_3$ = x – 15  when x = 39
   $\oplus_{3,3,8}$ (x, y) : $z_3$ = x – 16  when x = 42
   $\oplus_{3,3,9}$ (x, y) : $z_3$ = x – 18  when x = 43

*chatva sandhi*

4. *śaścho'ṭi* || 8.4.63 ||
   Rule: non-nasal mutes followed by *ś* which is in turn followed by a vowel, aspirate or *y, v, r* -> *ś* replaced by *ch*.
   General operator: $\oplus_{3,4}$ (x, y) = z = $xz_3$   when 24 <= x <= 43, y = 44, 1 <= w <= 17
   $\oplus_{3,4,1}$ (x, y) : $z_3$ = 36

### 4.4 $C_4$ *Sandhi*-s

*tugāgama sandhi*

1. *che ca* || 6.1.73 ||
   *āṅmāṅośca* || 6.1.74 ||
   *dīrghāt* || 6.1.75 ||
   *padāntādvā* || 6.1.76 ||
   Rule: vowel followed by *ch* -> *t* added.
   General operator: $\oplus_{4,1}$ (x, y) = z = $xz_4$y  when x <= 13, y = 36
   $\oplus_{4,1,1}$ (x, y) : $z_4$ = 41

*dhuḍāgama sandhi*

2. *ḍaḥ si dhuṭ* || 8.3.29 ||
   *naśca* || 8.3.30 ||
   Rule: *ḍ* or *n* followed by *s* -> *dh* added.
   General operator: $\oplus_{4,2}(x, y) = z = xz_4y$ when $x \in \{23, 32\}$, $y = 46$
   $\oplus_{4,2,1}(x, y) : z_4 = 28$

*tugāgama sandhi*

3. *śi tuk* || 8.3.31 ||
   Rule: *n* followed by *ś* -> *t* added.
   General operator: $\oplus_{4,3}(x, y) = z = xz_4y$ when $x = 23$, $y = 44$
   $\oplus_{4,3,1}(x, y) : z_4 = 41$

*ṅamuḍāgama sandhi*

4. *ṅamo hrasvādaci ṅamuṇṇityam* || 8.3.32 ||
   Rule: Short vowel precedes *ṅ*, *ṇ*, *n* which is followed by vowel -> *ṅ*, *ṇ*, *n* get duplicated.
   General operator: $\oplus_{4,4}(x, y) = z = xz_4y$ when $x \in \{21, 22, 23\}$, $1 \leq y \leq 13$, $u \in \{1, 3, 5, 7, 9\}$
   $\oplus_{4,4,1}(x, y) : z_4 = x$

4.5 $C_5$ *Sandhi*-s

*svādi sandhi*

1. *etattadoḥ sulopo'koranañsamāse hali* || 6.1.132 ||
   Rule: word *eṣaḥ* or *saḥ* followed by a consonant -> *visarga* (end *ḥ*) of first word dropped.
   General operator: $\oplus_{5,1}(x, y) = z = y$ when $x = 49$, $14 \leq y \leq 47$, $X \in \{10+45+1+49, 46+1+49\}$
   $\oplus_{5,1,1}(x, y) = z = y$

2. *so'ci lope cetpādapūraṇam* || 6.1.134 ||
   Rule: word *saḥ* followed by a vowel -> the final *visarga* of first word optionally dropped.
   General operator: $\oplus_{5,2}(x, y) = z = y$ when $x = 49$, $1 \leq y \leq 13$, $X = 46+1+49$
   $\oplus_{5,2,1}(x, y) = z = y$

3. *lopaḥ śākalyasya* || 8.3.19 ||
   Rule: final *v* or *y* preceded by *a* or *ā* and followed by a vowel, semi-vowel, nasal, column 3 or column 4 -> the *v* or *y* is dropped.
   General operator: $\oplus_{5,3}(x, y) = z = y$ when $x \in \{15, 16\}$, $1 \leq y \leq 33$, $u \in \{1, 2\}$
   $\oplus_{5,3,1}(x, y) = z = y$

4. *oto gārgyasya* || 8.3.20 ||
   Rule: *y* preceded by *o* and followed by a vowel, semi-vowel, nasal, column 3 or column 4 -> the *y* is dropped.
   General operator: $\oplus_{5,4}(x, y) = z = y$ when $x = 15$, $1 \leq y \leq 33$, $u = 11$
   $\oplus_{5,4,1}(x, y) = z = y$

5. *hali sarveṣām* || 8.3.22 ||
   Rule: *y* followed by consonant -> *y* dropped.
   General operator: $\oplus_{5,5}(x, y) = z = y$ when $x = 15$, $14 \leq y \leq 46$
   $\oplus_{5,5,1}(x, y) = z = y$

The aphorisms presented above encompass four out of the five major *sandhi* divisions that exist in Sanskrit as per Pāṇini – vowel, consonant, *prakṛtibhāva* (no change and hence not dealt with here), *visarga* and *svādi*. The vowel *sandhi*-s have been extensively dealt with above, with all exceptions to main rules incorporated. In the other divisions, only the main *sandhi*-s have been covered. Furthermore, listing of the same aphorism twice was necessitated by the need for different general operators for different rules within the same aphorism.

It is noteworthy that the *sandhi*-s have not been presented under these five divisions, but in the order of the categories introduced in this paper. Furthermore, since the order of aphorisms is crucial to determining the sequence of firing of the rules, Pāṇini's numbering (given as aphorism number for each aphorism) has been maintained, albeit only within each category.

## 5. Conclusions

In spite of there being almost 2500 individual letter-level rules (Table 2), this new schema that directly maps the patterning in the Pāṇinian aphorisms in a simple and effective way, ensures that we arrive at a total of just 110 equations. Clearly, this is a computationally lean way of calculating the result of *sandhi* operations. The results represent a computational model to process a majority of the euphonic conjunctions in Sanskrit. The work also demonstrates the simplicity with which euphonic conjunctions can be handled by adopting Pāṇini's precise scheme for rule representation.

A main strength of this modeling approach is that it is deterministic, as against the probabilistic methods adopted till now for *sandhi* operations. Determinism is inherent in Pāṇini's *sandhi* rules, which indeed specify how *sandhi*-s are formed and not how they are broken up, and this determinism has been uniquely tapped and modeled in this work. Traditional AI methods such as hidden Markov models, which have hitherto been applied for Sanskrit processing [3-7], assume relevance in the *sandhi*-splitting approach in which there are inherent ambiguities, rather than in the *sandhi*-building approach which is modeled here.

The five main operators and all the 110 derived equations designed and presented in this work, form the immediate basis for directly realizing crucial applications of *sandhi*-processing such as subtext searching.

**Rama N.** B.Sc. (Mathematics), Master of Computer Applications, Ph.D. (Computer Science). Employment: Anna Adarsh College, Chennai; Bharathi Women's College, Chennai; Presidency College, Chennai, India. Has 20 years of teaching experience including 10 years for PG, has guided 15 M.Phil. students; Chairperson of the Board of Studies in Computer Science for UG, and Member, Board of Studies in Computer Science for PG and Research at the University of Madras. Current research interest: Program Security. Member of Editorial cum Advisory Board of Oriental Journal of Computer Science and Technology.

**Meenakshi Lakshmanan.** B.Sc. (Mathematics), Master of Computer Applications, M.Phil. (Computer Science), currently pursuing Ph.D. (Computer Science) at Mother Teresa Women's University, Kodaikanal, India and pursuing Level 4 Sanskrit (*Samartha*) of the *Saṃskṛta Bhāṣā Pracāriṇī Sabhā*, Chittoor, India. Employment: SRA Systems Pvt. Ltd., currently Head of the Department, Department of Computer Science, Meenakshi College for Women, Chennai, India. Is a professional member of the ACM and IEEE.